\PassOptionsToPackage{unicode}{hyperref}
\PassOptionsToPackage{hyphens}{url}
\documentclass[
  11pt,
]{article}
\usepackage{xcolor}
\usepackage[margin=1in]{geometry}
\IfFileExists{upquote.sty}{\usepackage{upquote}}{}
\IfFileExists{microtype.sty}{
  \usepackage[]{microtype}
  \UseMicrotypeSet[protrusion]{basicmath} 
}{}
\makeatletter
\@ifundefined{KOMAClassName}{
  \IfFileExists{parskip.sty}{%
    \usepackage{parskip}
  }{
    \setlength{\parindent}{0pt}
    \setlength{\parskip}{6pt plus 2pt minus 1pt}}
}{
  \KOMAoptions{parskip=half}}
\makeatother
\usepackage{longtable,booktabs,array}
\usepackage{calc} 
\usepackage{etoolbox}
\makeatletter
\patchcmd\longtable{\par}{\if@noskipsec\mbox{}\fi\par}{}{}
\makeatother
\IfFileExists{footnotehyper.sty}{\usepackage{footnotehyper}}{\usepackage{footnote}}
\makesavenoteenv{longtable}
\providecommand{\tightlist}{%
  \setlength{\itemsep}{0pt}\setlength{\parskip}{0pt}}
\usepackage{amsmath,amssymb,amsthm}
\usepackage{graphicx}
\IfFileExists{xurl.sty}{\usepackage{xurl}}{}
\usepackage{hyperref}
\usepackage{bookmark}
\hypersetup{
  pdftitle={ReadingMachine: A Computational Methodology for Structured
Corpus Reading and Large-Scale Synthesis},
  pdfauthor={James Morrissey},
  colorlinks=true,
  linkcolor=black,
  citecolor=black,
  urlcolor=cyan
}

\title{ReadingMachine: A Computational Methodology for Structured
Corpus Reading and Large-Scale Synthesis}
\author{James Morrissey}
\date{\today}

\begin{document}
\maketitle

\section*{Abstract}\label{abstract}

The volume of written material produced across modern institutions now 
far exceeds the capacity of human analysts to read, interpret, and synthesize it comprehensively. Existing computational approaches—such as 
retrieval-augmented generation, hierarchical summarization, and agentic research 
workflows—scale access to information but typically operate over selectively 
accessed material, introducing risks of omission, early information loss, and 
limited analytical transparency. As a result, high-stakes synthesis tasks in 
research, policy, law, and organizational analysis remain constrained by 
partial coverage and limited inspectability.

This paper introduces ReadingMachine, a computational methodology for 
structured corpus reading that reframes large language models as systems 
for executing bounded reading operations alongside their use as generators 
of answers. Rather than producing synthesis in a single step, the method 
decomposes large-scale reading into a sequence of constrained, inspectable 
tasks—including insight extraction, semantic clustering, thematic schema 
construction, and iterative omission detection—prior to synthesis. By delaying 
irreversible compression and explicitly tracking intermediate representations, 
the approach prioritizes coverage, traceability, and the preservation of 
disagreement across large corpora. The approach is characterized by several design 
choices, including insight-level representation as the unit of synthesis, the use 
of clustering as scaffolding rather than determinant of themes, and explicit 
omission detection through orphan tracking, which together enable coverage-oriented, 
self-correcting synthesis under scale. The result is not synthesized output reading 
for general consumption, but instead an intermediate reading layer to aid with 
downstream decision-making.

The system is demonstrated on a heterogeneous corpus of 152 documents on industrial 
policy, producing over 17,500 atomic insights and a structured thematic mapping. We 
report system costs, runtime, qualitative characteristics of the output, and observed 
failure modes, including challenges related to synthesis under scale and omission 
detection. While formal comparative evaluation remains future work, a descriptive 
account of the results illustrates how structured corpus reading produces outputs 
that are substantively and epistemically distinct from conventional AI-based summarization.

ReadingMachine is released as an \href{https://github.com/morrisseyj/ReadingMachine/tree/main}
{open-source}, experimental framework intended to 
support systematic qualitative synthesis and downstream, explicitly constrained 
reasoning over stable corpus representations. The paper positions structured corpus 
reading as a complementary mode of knowledge production, distinct from retrieval-based 
and agentic AI systems. The results should be understood as a demonstration of system 
behavior at scale rather than a validated assessment of performance. Collaboration 
on evaluation is actively encouraged.

\section{Introduction}\label{introduction}

The volume of written information produced across modern institutions
now far exceeds the capacity of any individual---or even coordinated
teams---to read, interpret, and synthesize it. In domains ranging from
academic research and public policy to organizational learning and legal
analysis, the limiting factor is no longer access to information, but
the ability to systematically engage with it. As a result, critical
insights are routinely overlooked, minority perspectives are lost, and
synthesized outputs reflect invisible omissions rather than the full
structure of the underlying corpus.

Today, this problem is managed primarily through human-led review.
Literature reviews, policy syntheses, and internal analyses rely on
expert readers to select, interpret, and compress large bodies of text
into coherent narratives. While indispensable, this approach is
inherently constrained. Human review is slow and costly, limiting the
size of corpora that can be feasibly analyzed. It is bounded by
attention, making comprehensive coverage difficult to achieve in
practice. And it is path-dependent: as reviewers read, their priors
evolve, shaping subsequent interpretation and selection in ways that are
rarely visible or reproducible. The resulting analyses are valuable, but
they are also partial, contingent, and difficult to systematically
audit.

Existing computational approaches do not resolve these limitations.
Retrieval-based systems surface fragments of text but leave most of the
corpus unexamined. Hierarchical summarization pipelines compress
information early, often flattening disagreement and discarding nuance.
Agentic research systems explore selectively, producing analyses that
depend heavily on intermediate decisions and paths taken \cite{he2025}. 
In all cases, omission is common and difficult to detect
\cite{laban2024,liu2024}.

This creates a fundamental constraint on knowledge work. As writing
scales, reading does not---at least not in a way that preserves
coverage, traceability, and consistency. The result is a growing
mismatch between what is available to be known and what is actually
incorporated into analysis. In high-stakes contexts, where decisions
depend on comprehensive and faithful interpretation of large bodies of
text, this gap introduces systematic risk.

ReadingMachine is designed to address this constraint directly. It
begins from a different premise: the core challenge is not only
reasoning over large corpora, but reading them.

Current applications of large language models have largely been
developed as systems for generating answers---compressing information
into outputs conditioned on queries. This framing has proven highly
effective across a wide range of tasks, particularly where rapid
synthesis, exploration, or question answering is required. However, it
places emphasis on producing outputs from selectively accessed material,
rather than systematically engaging with the full structure of a corpus.

For many use cases, this is entirely appropriate. But for tasks where
coverage, traceability, and the preservation of disagreement are
central, this framing leaves important aspects of the problem
under-addressed. The limiting factor in large-scale analysis is not only
the ability to generate answers, but the ability to systematically read
and organize the underlying material.

Under this view, large language models can be more productively
understood not only as general-purpose reasoners, but also as systems
capable of performing bounded reading operations at scale. The central
challenge is therefore not to produce a synthesis in a single step, but
to coordinate many constrained acts of reading in a way that preserves
structure and reduces omission.

The novelty of ReadingMachine lies not in improving model performance,
but in how these capabilities are organized. The system decomposes
large-scale corpus mapping into discrete, inspectable tasks---such as
extracting atomic insights, grouping them semantically, and synthesizing
them into themes---and coordinates language models to execute these
tasks in sequence. This approach shifts compression to later stages of
the pipeline and introduces explicit mechanisms to track, detect, and
reintegrate omitted material.

This reframing has several implications. First, the marginal cost of
reading---both in time and money---declines substantially, enabling
analysis over corpora that would be impractical to process through human
review alone. Second, by structuring reading as a sequence of explicit
transformations, the system enables corpus-level analysis that
emphasizes coverage (i.e.~prioritizes achieving the incorporation of all
relevant claims/arguments/insights from the corpus), traceability, and
the preservation of disagreement, while making intermediate steps
available for inspection. Third, by applying a computational approach,
the system meaningfully increases the scope for structural
reproducibility of qualitative data analysis. Fourth, separating reading
from downstream interpretation increases transparency and allows
reasoning to be applied more deliberately. Finally, because the
components of the pipeline can be varied independently, the system
supports controlled experimentation over how corpora are organized and
interpreted.

This paper introduces ReadingMachine as both a technical framework
and a methodological intervention. It outlines the architectural design
of the system, the analytical assumptions it encodes, and the trade-offs
it makes in shifting from selective, human-limited synthesis toward
structured, inspectable corpus mapping. More broadly, it situates the
approach within an information landscape where the central challenge is
no longer producing knowledge, but ensuring that it can be
systematically read, organized, and understood at scale.

The paper makes four contributions: (1) it introduces structured corpus
reading as a methodological framing for large-scale synthesis; (2) it
describes ReadingMachine, an open-source pipeline implementing this
approach; (3) it demonstrates the system on a 152-document industrial
policy corpus; and (4) it identifies observed scaling constraints and
failure modes, including orphan persistence, context-pressure effects,
and late-stage reinflation.

Because the system produces a corpus map rather than a query response 
or short summary, direct comparison with RAG or standard summarization 
systems is not straightforward: the relevant evaluation question is not 
only whether the final prose is better, but whether the intermediate 
representation improves coverage, traceability, disagreement preservation, 
and downstream usefulness.

ReadingMachine is released as an experimental method and open-source implementation
(\href{https://github.com/morrisseyj/ReadingMachine/}{GitHub repository}). It is
not presented as a finalized solution to large-scale synthesis, but as a
structured approach that can be inspected, modified, and evaluated
across different contexts and configurations. The properties described
in this paper---such as improved coverage, omission detection, and
traceability---should be understood as design objectives and observed
behaviors in a single large-scale run, rather than as formally validated
performance characteristics. Support on evaluation and testing of the
tool is actively encouraged.

\section{The Problem}\label{the-problem}

The production of written language has scaled dramatically over time.
From the printing press to digital publishing, networked communication,
and mass storage, the rate at which text is generated has increased by
orders of magnitude. Reading, however, has not undergone a comparable
transformation. While mass literacy expanded who can read, the act
itself remains constrained by attention, time, and sequential
comprehension. As a result, a structural imbalance has emerged: the
supply of text grows exponentially, while the capacity to process it
grows incrementally.

Large language models appear, at first glance, to resolve this
constraint. They can process and summarize text at speeds and volumes
far beyond human capability, and perform reliably on many comprehension
and synthesis tasks. However, current approaches to applying these
systems to large corpora inherit important limitations from their
underlying architectures \cite{narayan2018}.

Retrieval-augmented generation (RAG), the dominant paradigm, operates by
selecting subsets of a corpus in response to a query. Documents are
chunked, indexed, and retrieved based on estimated relevance, with only
the retrieved material entering the model's context \cite{lewis2020}.
This introduces a fundamental limitation: material that is not retrieved
is not analyzed, and the system has no visibility into what has been
excluded. As a result, omission is both common and difficult to detect.
As corpora grow, this problem is compounded by context constraints.
Increasing the number of retrieved passages places pressure on the
model's attention, leading to known degradation effects such as the
``missing middle,'' where portions of the input receive less effective
processing \cite{liu2024}, and dropped insights during
synthesis/integration \cite{laban2024}.

Hierarchical summarization pipelines take a different approach, but
encounter related issues. By compressing documents into summaries and
then recursively summarizing those summaries, these systems introduce
information loss early in the process. Minority or low-frequency claims
are often dropped, disagreements are smoothed, and the resulting
representations tend toward generalization at the expense of
specificity. Each stage of compression reduces the recoverability of the
original material \cite{cohan2018}. Problems of missing middle and
dropped insights during synthesis/integration repeat here as context
window pressure grows \cite{liu2024,laban2024}.

Agentic research workflows attempt to mitigate these issues through
iterative search and refinement. However, because they depend on
sequences of intermediate decisions---what to query, what to retrieve,
how to interpret---small variations in prompts or trajectories can
produce substantially different results \cite{yao2022,he2025}. 
Coverage remains partial, and omissions remain implicit.

For these reasons, high-stakes synthesis tasks over large corpora
continue to rely on human-led review. This approach provides judgment
and contextual interpretation, but it does not resolve the underlying
scaling problem. Human review is slow and expensive, limiting the
feasible size of a corpus. It is constrained by attention, making
comprehensive coverage difficult in practice. And it is path-dependent:
as reviewers read, their priors evolve, shaping subsequent
interpretation and selection in ways that are not fully observable or
reproducible.

The result is a persistent gap between what is written and what is
systematically understood. In contexts where omission is costly---such
as policy analysis, legal review, or scientific synthesis---this gap
introduces a form of structural risk. The challenge is not simply to
access or summarize information, but to read large corpora in a way that
preserves coverage, traceability, and the internal structure of
arguments.

\section{ReadingMachine: Structured Corpus
Reading}\label{readingmachine-structured-corpus-reading}

To address these limitations, ReadingMachine proposes a different
approach to large-scale corpus analysis. Rather than retrieving subsets
of documents or compressing texts through successive summaries, the
system performs a structured reading pass across the \textbf{entire}
corpus, extracting and organizing atomic insights prior to synthesis.

The core design principle is to separate reading from interpretation.
This distinction is important. Many contemporary applications treat
language models as general-purpose reasoning systems, capable of
synthesizing and evaluating complex bodies of information in a single
pass. ReadingMachine adopts a different approach: it decomposes the act
of reading into bounded, inspectable tasks---such as insight extraction,
semantic cluster summarization, and theme formation. This constrains the
model to perform narrowly defined reading tasks over bounded inputs, and
defers interpretation, evaluation, and judgment to the researcher or
downstream analysis. The system is therefore not designed to ``reason
over'' a corpus, or to determine which claims are correct, important or
true in a corpus, but to produce a structured representation of what the
corpus contains. At the same time, this approach preserves traceability
between synthesized outputs and their source material.

The architecture also implements explicit mechanisms to ensure coverage
and to preserve conflict during synthesis. This structure is intended to
reduce omission, maintain granularity, and make the analytical process
inspectable and reproducible.

\section{Methodology}\label{methodology}

ReadingMachine operates over a wide range of natural-language corpora,
including academic literature, policy reports, organizational documents,
legal materials, and qualitative data. It is designed for structured
synthesis tasks over a defined corpus.

ReadingMachine is not a systematic review method: it does not assess
study quality, weight evidence, or produce adjudicated conclusions. Its
purpose is corpus mapping---structuring what a defined body of text
contains. In this respect the method is closer to a ``scoping study'' as
articulated by Arksey and O'Malley \cite{arksey2005}. ReadingMachine achieves this
through a computational approach that coordinates structured reading
passes over the entire corpus to produce a thematic map of its contents
without adjudicating evidentiary quality.

At the same time, the pipeline draws on the logic of reflexive thematic
analysis \cite{braun2006}, particularly the progression from
coding to theme development, review, and refinement. Extracted insights
function as computational analogues of codes, while theme construction
and iteration are implemented as explicit, inspectable stages.

In abstract form, the workflow can be understood as:

reading a corpus $\rightarrow$ extracting insights $\rightarrow$ 
organizing them into themes $\rightarrow$ synthesizing $\rightarrow$ 
checking for completeness $\rightarrow$ refining structure

This process is implemented as a sequence of constrained, inspectable
transformations. Each stage isolates a specific analytical
function---such as insight extraction, clustering, or thematic
synthesis---so that intermediate representations remain visible and
omissions can be detected and corrected. Interpretive judgments are not
removed, but they are externalized and structured, allowing them to be
inspected and compared across runs.

The resulting pipeline is shown below in Figure~\ref{fig:pipeline}.

\begin{figure}
\centering
\includegraphics[width=0.85\textwidth]{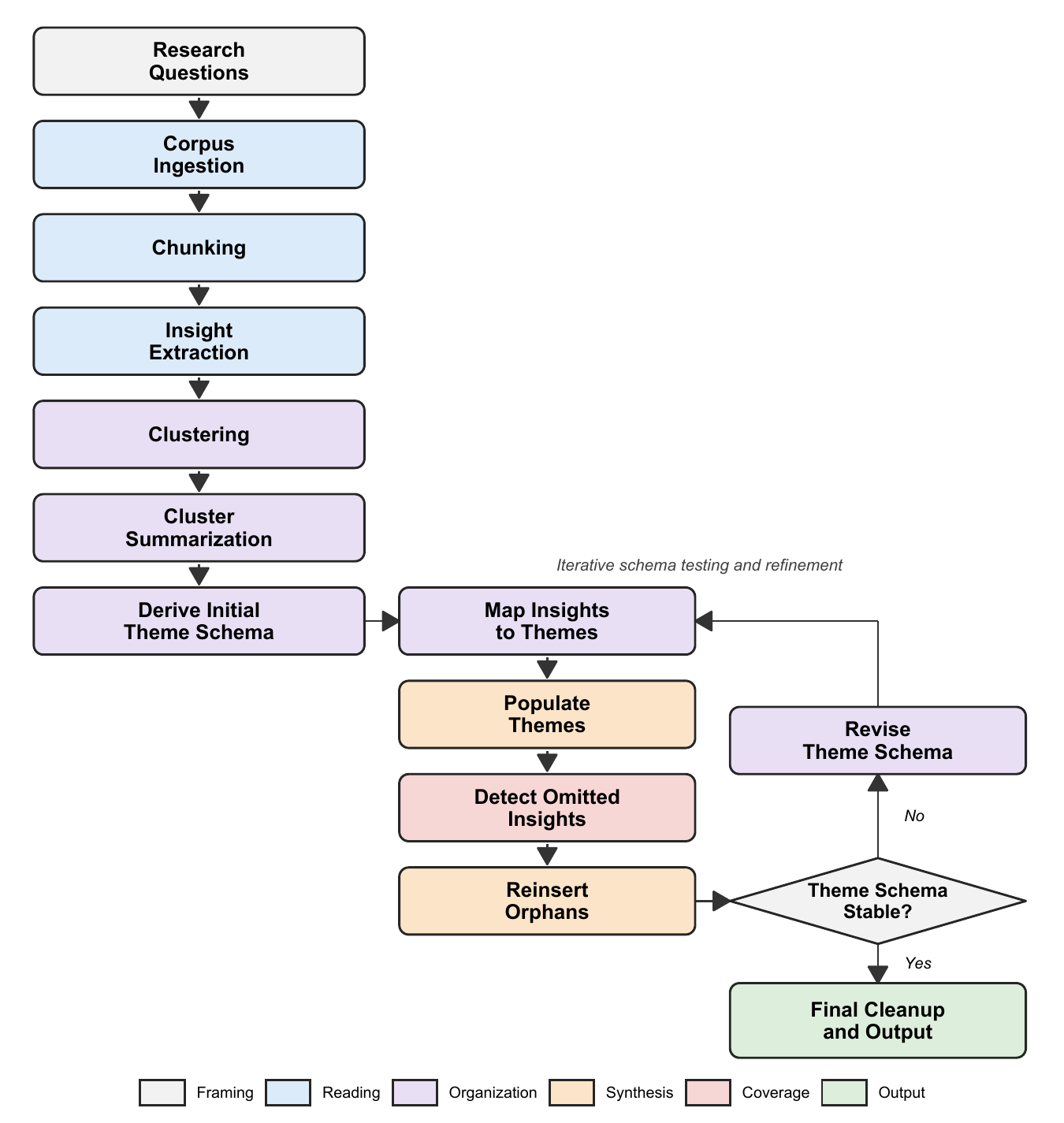}
\caption{The system decomposes corpus analysis into structured reading 
stages, extracting atomic insights prior to synthesis. Clustering provides 
semantic scaffolding for theme generation, while synthesis occurs at the 
theme level. An explicit orphan detection and reinsertion mechanism enforces 
coverage, and an iterative loop refines the thematic schema until stable.}
\label{fig:pipeline}
\end{figure}

The system is implemented, in Python, as a sequential pipeline in which
each stage transforms a shared state object. Analytical choices---such
as the definition of insights, extraction instructions, clustering
parameters, and theme construction rules---are fixed, inspectable, and
version-controlled, making the structure of the synthesis explicit
rather than embedded in a single opaque step.

\subsection{Generate Research
Questions}\label{generate-research-questions}

Prior to processing the corpus, the user defines a set of research
questions that guide the analysis. These questions determine what
constitutes a relevant insight and shape all subsequent stages of the
pipeline. ReadingMachine does not generate or refine these questions;
they reflect the user's analytical intent and frame the scope of the
synthesis.

\subsection{Ingest and Chunk the
Corpus}\label{ingest-and-chunk-the-corpus}

Documents are ingested into the system, with the current implementation
supporting PDF and HTML formats. During ingestion, basic metadata (e.g.,
title, author, year) is extracted using a language model.

The corpus is then divided into smaller semantic units through a
chunking process. Chunking uses a greedy, size-constrained strategy to
produce segments of approximately fixed length (e.g.,
\textasciitilde3500 characters), ensuring consistent coverage of the
source text. Within each segment, the splitter preferentially backtracks
to the nearest paragraph break, sentence boundary, or whitespace to
avoid arbitrary truncation where possible. Preserving strict size
constraints takes priority over perfect semantic boundaries.

\subsection{Generate Insights}\label{generate-insights}

Each chunk is processed alongside the full set of research questions.
The language model is instructed to extract discrete insights---defined
as atomic claims, arguments, or findings in the text that are relevant
to the questions. Each insight is returned as a concise, one- to
two-sentence statement. ``Atomic'' here refers to the smallest
operational unit of representation within the system, not to a
non-interpretive or pre-analytic unit of meaning.

Insights are formally defined in the prompt as:

\begin{quote}
``any explicit claims, arguments, findings, or statements in the text
that bear on {[}the{]} question{[}s{]}\ldots{} An explicit claim
includes: stated findings or conclusions, causal statements (e.g.~X
leads to Y), explanations of mechanisms or processes, descriptive
statements that clearly assert a relationship, condition or effect. They
are not restricted to formal conclusions {[}as{]} many valid claims
appear as descriptive or explanatory statements. Each {[}insight{]} must
be concise (one sentence or short phrase) and preserve wording as much
as possible.''
\end{quote}

To complement this chunk-level pass, the system performs a second,
document-level pass. In this stage, the full document (or the largest
portion that fits within the model's context window) is processed with
each research question individually. The model is provided with the
previously extracted chunk-level insights and instructed to identify
additional insights without repetition.

Insights are distinguished from meta-insights as follows in the prompt:

\begin{quote}
``{[}meta-insights are{]} higher-level, traceable arguments or
conclusions that span across multiple parts of a text \ldots{}
Chunk-level insights capture localized claims. {[}meta-{]}insights
\ldots{} ONLY become visible when combining information across multiple
parts of the document. {[}meta-insights{]} must introduce a
substantively new claim that only becomes visible when considering
multiple parts of the document together''
\end{quote}

This dual approach captures both localized claims and broader arguments
that span multiple sections of a document, reducing the likelihood that
cross-cutting ideas are missed during chunk-level processing.

\subsection{Cluster Insights}\label{cluster-insights}

Extracted insights are organized using unsupervised learning techniques
to group semantically similar claims. Clustering is not a replacement
for a conceptual mapping, but serves as scaffolding to the process of
conceptual theme generation---see below. This is necessary to afford the
model a structured representation of the corpus, upon which to build
themes.

Clustering proceeds in a fashion similar to that used in BERTopic
\cite{grootendorst2022}. Each insight is first converted into a vector
representation (embedding). Because these embeddings exist in a
high-dimensional space, dimensionality reduction is applied using UMAP,
due to its speed and flexibility \cite{mcinnes2018}, to improve
clusterability and mitigate sparsity effects associated with
high-dimensional data.

Clustering is then performed using HDBSCAN, due to its robustness to
parameter selection and effective density-based approach \cite{mcinnes2017}, which groups insights based on density while allowing for the
identification of outliers. The system includes parameter sweep
functionality to support selection of appropriate configurations. UMAP
parameters are evaluated using silhouette scores, with research
questions used as reference labels to assess separation (the user can
exclude subsets of research questions in cases of conceptually
overlapping questions). HDBSCAN configurations are assessed using a
combination of Davies--Bouldin scores and outlier minimization. Note
that while clustering generates the semantic surface that theme
generation operates over (see below), it does not fundamentally
determine themes. Themes are instead iteratively refined via an orphan
reinsertion and re-theming loop (see below).

Dimensionality reduction is applied globally across the insight set,
while clustering is performed separately for each research question.
From this stage onward, all subsequent steps operate on a per-question
basis. The final output is therefore composed of parallel syntheses for
each research question.

This design introduces some redundancy across the overall report, as
similar insights may appear in multiple question-specific outputs. This
is intentional: each question is treated as a self-contained analytical
unit, allowing users to engage with individual sections independently
without requiring full traversal of the entire synthesis. This is a
common reading style for literature synthesis.

\subsection{Theme Generation}\label{theme-generation}

Clusters group insights by semantic similarity, providing a useful
approximation of structure within the corpus. However, semantic
proximity does not necessarily correspond to conceptual organization.
This distinction is important methodologically. In Braun and Clarke's account \cite{braun2006}, themes are not simply discovered as pre-existing
clusters in the data; they are actively constructed through analytic
judgment. ReadingMachine preserves this distinction by using clustering
only as scaffolding for thematic interpretation, rather than treating
clusters themselves as themes. To move from semantic groupings to
analytically meaningful categories, the system generates themes through
a two-stage process.

First, each cluster of insights is summarized. Clusters are processed in
an order determined by the shortest path between cluster centroids in
the embedding space. This ordering places semantically similar clusters
adjacent to one another. As each cluster is summarized, the last five
generated summaries are provided as frozen context for subsequent steps.
This allows for local coherence without risking anchoring on initial
clusters summaries (which could happen if global context was frozen).
All clusters, including outliers, are included in this process.

The result is a sequence of cluster summaries that forms a structured
narrative of the corpus, organized by semantic proximity and with
reduced repetition. This narrative serves as the input for theme
generation, allowing the model to operate over an ordered representation
of the corpus rather than a disorganized set of individual insights.

Second, the cluster summary narrative is passed to a language model to
generate a theme schema. This schema defines a set of thematic
categories (theme label and theme description), along with rules for
assigning insights to each theme. The schema includes an \emph{Other}
category to capture minority insights that do not warrant a dedicated
theme, preventing unnecessary proliferation of categories. This category
is distinct from clustering outliers; outlier insights may be assigned
to any theme based on their content.

The schema also includes a \emph{Conflicts} category when substantively
incompatible claims are present. This is introduced explicitly to
counter the tendency of language models to smooth disagreement during
summarization and to preserve contested areas of the corpus. Conflict is
defined in the prompt as:

\begin{quote}
``claims, or prescriptions that cannot be jointly maintained within a
single coherent analytical frame. {[}Conflicts are not{]}: multiple
reinforcing critiques, layered constraints or complexities, {[}the
articulation of{]} trade-offs within a shared analytical orientation,
{[}expressions of{]} variations that do not represent incompatible
positions. Conflicts \ldots{} {[}require{]} identifiable polarity
between positions''
\end{quote}

The output of this stage is a theme schema that represents a conceptual
mapping of the corpus. Themes function as structured containers for
insights, enabling synthesis at the level of arguments rather than
documents.

\subsection{Theme Mapping and
Population}\label{theme-mapping-and-population}

Once a theme schema has been defined, insights are assigned to themes.
Insights are processed in batches alongside the schema, with the
language model instructed to map each insight to one or more relevant
themes based on the inclusion and exclusion rules generated in the
schema development pass. A single insight may be associated with
multiple themes where appropriate. The Other category is populated both
according to its defined criteria and as a fallback for insights that do
not clearly fit within existing themes. Additional controls are applied
to ensure that the language model only returns valid insight identifiers
for each batch.

After all insights have been mapped, each theme---now represented as a
theme label, theme description, and collection of associated
insights---is passed to the language model for summarization. Because
synthesis operates directly over structured insight sets rather than
full documents or intermediate summaries, compression occurs at a late
stage in the pipeline. Each theme is generated independently, using only
the insights assigned to it, which allows the model to focus on a
bounded and conceptually coherent subset of the corpus during each
synthesis step.

It is worth noting that compression does occur at multiple stages of the
pipeline, prior to summarization: during clustering (dimensionality
reduction) and schema generation. However, these compressive acts are
used to organize the insight space. They are not destructive compressive
acts in that they do not replace or discard the underlying insight set.
Irreversible compression occurs only at the point of theme-level
synthesis, which operates directly over the full set of assigned
insights, and even then raw insights are reinserted in the orphan
handling loop---see below.

\subsection{Orphan Handling}\label{orphan-handling}

To reduce omission risk during theme-level synthesis, the pipeline
explicitly identifies and reintegrates orphan insights. Orphans
\textbf{are NOT clustering outliers or insights assigned to the Other
category}. Rather, they are insights that were mapped to a theme but do
not appear in that theme's synthesized summary. As such, they represent
omissions introduced during the theme synthesis stage. This design
responds to known long-context failure modes, including position
sensitivity and incomplete coverage of relevant insights during
summarization \cite{liu2024,laban2024}.

Orphan detection is performed by comparing the set of insights assigned
to each theme with the content of the corresponding theme summary.
Insights are processed in batches alongside the summary, and the
language model is instructed to identify which insights are
substantively represented. Any assigned insights not identified in this
process are classified as orphans. The Orphan identification prompt is
intentionally strict to drive false positives, as a guard against
omission.

Once identified, orphan insights are reintroduced, with orphans and the
current theme passed to a language model with instructions to insert
them into the existing theme, and return a full re-write. Orphan
handling therefore is a mechanism for pursuing coverage at the theme
level. This serves to reduce the likelihood of silent omission during
synthesis. Orphan insertion also plays a central role in theme iteration
(see below)

\subsection{Theme Iteration}\label{theme-iteration}

The initial theme schema is generated from cluster summaries, which
introduces an early stage of compression. As a result, the initial set
of themes may not fully capture the granularity or range of insights
present in the corpus. After orphan handling, all insights have been
forcibly incorporated into the existing themes, which can strain or
destabilize the initial thematic structure.

To address this, the pipeline regenerates the theme schema using the
current theme summaries as input. This second pass operates over a
representation that now includes all insights, allowing the model to
refine, split, or reorganize themes to better reflect the underlying
material. Following schema regeneration, insights are remapped to
themes, summaries are regenerated, and orphan handling is applied again.

This theme generation $\rightarrow$ mapping $\rightarrow$ orphan handling loop is repeated
iteratively. This iterative review process parallels the theme review
and refinement phases described by Braun and Clarke \cite{braun2006}, in which
candidate themes are checked against the underlying data and revised
where they fail to adequately capture the dataset. In ReadingMachine,
orphan detection operationalizes this by identifying insights that have
not been substantively represented in the current thematic synthesis. In
the current implementation the model is instructed not to make
speculative improvements to the schema and only offer an updated schema
when the input material shows obvious opportunities for improved
structuring. In this respect an ideal schema prioritizes:

\begin{quote}
\begin{enumerate}
\def\labelenumi{\arabic{enumi}.}
\tightlist
\item
  Conceptual coherence within each theme
\item
  Clear conceptual boundaries between themes
\item
  Complete conceptual coverage of the data
\item
  Minimizing reliance on the `Other' category only if the above are
  satisfied
\item
  Minimizing the number of themes only if the above are satisfied
\end{enumerate}
\end{quote}

If the model cannot see obvious improvements for a question's themes,
they are all marked as stable. Only unstable themes are passed to future
iterations. Once all themes are stable the user is instructed to move to
redundancy handling and summarization. In future it likely makes sense
to provide some formal measure of schema stability to track how it
evolves over time considering: stability of themes, size of Other
category, stability of insight allocation to themes, number of orphans
etc.

\subsection{Redundancy Pass}\label{redundancy-pass}

Because themes are generated and populated independently, and because
individual insights may be assigned to multiple themes, redundancy
across theme summaries is expected. To address this, the system includes
an optional redundancy reduction pass.

In this stage, theme summaries for a given research question are
processed sequentially, with previously generated summaries provided as
context. The language model is instructed to reduce repetition by
introducing cross-references (e.g., ``as discussed above'') while
preserving all substantive content.

This step improves readability but introduces a risk of information
loss, particularly for insights that were marginal or recently
reinserted during orphan handling. For this reason, the redundancy pass
is optional. Users can choose between a completeness-first output, which
retains some repetition, and a readability-first output, which reduces
redundancy at the potential cost of minor omissions.

\subsection{Rendering}\label{rendering}

The final stage of the pipeline produces a structured output that can be
rendered as a report. This includes optional generation of a title,
executive summary, and research question--level summaries, each created
through targeted language model calls.

An additional optional step applies a stylistic rewrite to improve
readability. In this pass, theme summaries are processed sequentially,
with prior sections provided as context, and the model is instructed to
vary phrasing and reduce repetitive or mechanical language. The goal is
to produce a more natural and readable document without altering the
underlying content.

As with the redundancy pass, this step introduces a trade-off. While it
improves readability, it carries a risk of minor information loss or
unintended modification of phrasing. It is therefore optional, allowing
users to choose between a more literal, completeness-preserving output
and a more polished, reader-friendly version.

The final output can be rendered as a PDF, markdown, or Word document.

\section{Scaling the methodology}\label{scaling-the-methodology}

The primary scaling constraint for any LLM is the context window, which
comprises both input and output tokens. In addition to this hard limit,
there are softer constraints related to attention degradation and
``missing middle'' effects, which arise when the context window becomes
large or contains complex content. The methodology described above is
designed primarily to mitigate these soft limitations by constraining
the load placed on the model: bounding reading tasks, batching
operations, explicitly identifying and reinserting omissions, and
iterating on thematic structure.

However, when applied to large and heterogeneous corpora, the method can
encounter hard context window limits at specific stages of the pipeline.
These limits were observed in practice during large-scale execution. In
particular:

\begin{itemize}
\tightlist
\item
  too many insights passed during cluster summarization (input window
  exceeded)
\item
  too many insights passed during theme population (input window
  exceeded)
\item
  too many orphans passed during reinsertion (input window exceeded)
\item
  the requirement to represent all orphans while maintaining
  insight-level granularity during reinsertion can exceed the output
  window, resulting in truncated outputs
\end{itemize}

These constraints do not invalidate the methodology, but they do affect
how its core properties---coverage, granularity, and structured
synthesis---can be maintained under finite context limits. When
intermediate representations cannot be fully processed, downstream
stages operate on partial inputs, increasing orphan counts and placing
additional pressure on reintegration steps. If unaddressed, this can
propagate through the pipeline and degrade the system's ability to
maintain a complete and structured representation of the corpus.

To preserve the methodological commitments of the system under these
constraints, a set of modifications were introduced. These modifications
do not alter the underlying logic of the pipeline, but adapt its
execution to operate within bounded context windows while maintaining
coverage through iterative reintegration and refinement.

The following sections describe these adjustments.

\subsection{Too many insights per
cluster}\label{too-many-insights-per-cluster}

To address cases where clusters contain more insights than can be
processed within the context window, a secondary, density-seeded
partitioning method is introduced. This method subdivides
clusters---including outliers---into smaller, semantically coherent
partitions that fall within the model's input constraints, while
preserving the underlying structure of the insight space.

The initial clustering step proceeds as described above, using UMAP for
dimensionality reduction and HDBSCAN for density-based clustering. For
any cluster exceeding a predefined size threshold (k), the following
partitioning procedure is applied:

\begin{itemize}
\tightlist
\item
  Calculate the required number of partitions: ceil(cluster size / k)
\item
  Use HDBSCAN to identify the single densest point in the cluster and
  assign it as a seed
\item
  Identify the nearest neighbors to that seed, up to k points and remove
  these points from the data pool
\item
  Repeat steps 2 and 3 until all partition seeds have been initialized
\item
  Return all data points to the pool, along with the identified seeds
\item
  Allocate points to partitions using a round-robin assignment based on
  nearest-neighbor distance to each seed, removing each point once
  assigned
\item
  Continue until all data points have been allocated
\end{itemize}

This process produces a set of semantically meaningful sub-clusters,
each of size \textless{} k, including those derived from outlier
regions. These partitions are then ordered by shortest distance between
their centroids and summarized using the frozen-context procedure
described above.

This adjustment allows cluster-level summarization to proceed within
context limits while preserving local semantic structure. During
implementation k is set smaller for outliers than it is for HDBSCAN
clusters, to account for the greater heterogeneity of the former.

\subsection{Too many insights passed to a theme during
population}\label{too-many-insights-passed-to-a-theme-during-population}

In cases where the number of insights mapped to a theme exceeds the
model's input capacity, a sampling strategy is introduced to ensure that
theme population can proceed within context limits. Specifically,
insights are greedily and randomly sampled up to a fixed threshold
(e.g., \textasciitilde70,000 words), ensuring that the input window is
not exceeded.

This introduces partial visibility at the point of theme synthesis.
However, this does not undermine the overall objective of coverage. The
architecture is designed to recover omitted material through the orphan
identification and reinsertion process, combined with iterative theme
schema refinement. In this way, sampling acts as a temporary constraint
on individual synthesis steps, while coverage is pursued across
iterations of the pipeline.

This adjustment allows theme population to operate within bounded input
constraints while maintaining the broader integrity of the method.

\subsection{Too many orphans passed to reinsertion loop}\label{too-many-orphans-passed-to-reinsertion-loop}

In cases where the number of orphan insights exceeds the model's context
window, orphan reinsertion is performed in batches. Each batch of
orphans is processed sequentially, with the full theme summary rewritten
at each step.

This approach increases computational cost, as the entire theme output
is regenerated for each batch. However, given the central role of orphan
reinsertion in maintaining coverage and refining the thematic structure,
this trade-off is necessary. Batching allows the reintegration process
to proceed within context constraints while preserving the system's
ability to iteratively incorporate omitted material.

\subsection{Output constraints and synthesis under bounded
length}\label{output-constraints-and-synthesis-under-bounded-length}

A further constraint arises from the requirement to represent all orphan
insights while maintaining insight-level granularity within a finite
output limit. During orphan reinsertion, this can cause the model to
exceed its maximum output tokens. When this occurs, outputs are
truncated and invalid JSON is returned.

This failure reflects a structural limit: it is not always possible to
fully represent all assigned insights at the required level of
granularity within a single bounded output. Rather than treating this
solely as an execution failure, the system uses it as a diagnostic
signal of excessive conceptual heterogeneity within a theme.

This is handled as follows:

\begin{enumerate}
\item When an orphan insertion batch fails:
  \begin{enumerate}
  \item A detailed summary of the failed batch is generated.
  \item The insertion process proceeds to subsequent batches.
  \item Once all batches have been processed, the failed batch summaries are appended to the theme summary.
  \item The theme is marked as incomplete; themes with no failed batches are marked as complete.
  \end{enumerate}

\item The resulting themes---now including summaries of failed insertions---are passed back to the theme 
schema generation step, with explicit instructions to prioritize resolving these failures. This is done 
primarily by splitting themes, and only by redistributing concepts where other themes have sufficient 
capacity, i.e., conceptual coherence and manageable size.

\item This process is repeated until all themes pass the completeness check.
\end{enumerate}

Under this design, orphan handling serves not only to address omission
and refine the thematic structure, but also to operationalize synthesis
under bounded output constraints.

While this orphan handling loop is effective, it also resulted in two 
further failure modes:
\begin{itemize}
  \item performative repair of the theme schema
  \item dropped citations during orphan reinsertion
\end{itemize}

Each of these were addressed as follows:

\subsubsection{Performative repair}\label{performative-repair}

Step 2, above, was initially attempted as a single LLM call. This did not 
work reliably: the schema repair pass repeatedly reproduced the same failing 
themes even when given explicit instructions to materially amend them. 
Providing additional context, including the full history of prior schemas, 
efforts at population, including failed repair attempts, did not resolve the 
issue. A self-reported repair accountability mechanism was then added, 
requiring the model to explain what repairs it had made. This also failed: 
the model sometimes retained the unchanged failing themes while reporting that 
repairs had been implemented.

This failure mode suggests that, under schema-repair pressure, the model could 
satisfy the rhetorical form of repair without performing the underlying structural 
transformation required to resolve the failure. One interpretation is that the 
model preferentially stabilized conceptually coherent schemas even when those 
schemas were operationally incompatible with bounded synthesis constraints.

An initial solution was to separate repair diagnosis from repair implementation. 
In the first pass, the model generated a schema repair plan based on the history 
of failed schemas and incomplete themes. In the second pass, a separate call 
implemented that plan against only the most recent schema, without access to 
the prior summaries or accountability narratives. The model priors towards producing 
conceptually coherent synthesis object however was so strong that the repair planner 
became too averse to splitting conceptually coherent themes even when the system was 
clearly being shown that they were conceptually overloaded and could not pass orphan 
reinsertion. 

The eventual solution was to split the repair pass first into a decomposition 
prompt where the sole aim was to break up failing themes into smaller unit, by 
taking out the largest coherent cluster of concepts that could stand alone as a 
theme without itself resulting in conceptual overload. This pass had to make no 
mention of conceptual elegance. Only after a theme was decomposed so that it would 
survive orphan insertion without errors was it passed it to an optimization prompt 
which was instructed to recombine coherent conceptual groups and clarify boundaries, 
so long as this did not risk recreating conceptual overload for any theme. After 
optimization themes are passed again through the orphan insertion loop to see if 
they remain viable. If the optimizer can see no obvious opportunities for improvement 
it defines the schema for the question as "stable". Once all question schema 
are stable, the iterative process completes. 

\subsubsection{Dropped citations}\label{dropped-citations}

The demands on orphan insertion (large numbers of orphans) eroded citation retention. 
Citations for papers that produced insights would drop out of the final summary 
altogether - such that around 1/3 of the total citations were missing from the final 
paper. This raises the possibility that underlying insights were also being lost during 
iterative synthesis, although the batched orphan architecture suggests citation erosion 
was more common than complete insight deletion. To handle this the following process was implemented:

\begin{enumerate}
\item After theme population an LLM call on the theme summary identifies all current citations
\item With each orphan batch the list of required citations is sent to the LLM, 
along with the batch of orphan insights to be incorporated. The instructions 
are to incorporate orphans and ensure that all the citations in the list appear in the theme. 
\item This is bootstrapped by a final citation check where the set of citations 
for the insights mapped to the theme is compared with the insights currently 
appearing in the theme. For any missing citations, all the insights for those 
citations for the theme, are then sent to an LLM with instructions to insert distinct citations at least once. 
\end{enumerate}

Notably this approach to citation repair did not initially work. Even though the 
model was provided with a full set of required citations and the relevant insights to 
insert, it would not successfully reinsert them. The hypothesized reason for this is 
similar to that explaining failed schema repair: the model is trained to output coherent 
readable paragraphs; these do not generally include large numbers of citations behind 
general claims. This preference persisted even when the model was explicitly instructed 
to stuff the citations. In testing, only about one third of the missing citations would 
get inserted into the text. To address this an approach similar to that used in schema 
repair was adopted, whereby instead of asking the model to both track insertion of missed 
citations and generate an updated complete synthesis, the model was tasked with generating 
a repair patch. This could instruct the addition of a citation to an existing claim or 
the addition of a new sentence to reflect the claim. The patch included, verbatim, the 
sentence to be replaced, or the preceding sentence (in the case of a new sentence). These were 
then deterministically replaced in python. The model occasionally errored in returning 
the exact sentence which is what is thought to account for the loss of citations. This 
problem should be fixable through the use of a fuzzy match on sentences or via a semantic 
match. This approach was not however implemented in this version of the tool.

The issue of dropped citations raises a question about the role we expect citations to 
play in the synthesis. It would be technically possible to have every claim list every 
citation that supports it in the paper, but this would undermine readability and erode 
token budgets where the hard constraint on operationalizing the pipeline is bounded output 
length. A compromise was therefore adopted that looks something more like human review. 
All claims should be cited, with no claim needing more than four citations, prioritizing 
the most prominent. At the same time, all authors that produce insights for a theme should 
appear at least once in the theme. In the final output this is partially achieved. The theme 
summary does not invoke more than four citations to support a point, but the citation repair 
can insert as many as it likes. The result is some claims supported by large numbers 
(e.g. 14) of citations. This slightly undermines readability, but for an intermediate 
reading layer this is considered acceptable. 

\section{Methodological Contributions}\label{methodological-contributions}

Within this method the following novel methodological contributions to
large‑scale qualitative corpus analysis are worth highlighting:

\begin{itemize}
\tightlist
\item
  \textbf{Insight‑level representation} as the atomic unit of synthesis,
  enabling late‑stage integration over granular claims rather than early
  document‑level compression.
\item
  \textbf{Formal semantic clustering as scaffolding for qualitative
  thematization}, with clustering used to expose structure rather than
  to determine themes.
\item
  \textbf{Scale‑aware synthesis techniques}, including shortest-path
  cluster ordering and density-seeded partitioning to preserve semantic
  neighborhoods under bounded context windows.
\item
  \textbf{Orphan‑based omission detection}, which treats synthesis gaps
  as explicit signals and uses them to iteratively refine thematic
  schemas under coverage constraints.
\item
 \textbf{External auditing of iterative LLM synthesis}, using explicit 
conservation checks for coverage, provenance, thematic coherence, and citation 
retention across repeated generative transformations
\end{itemize}

Individually, some of these components draw on existing techniques from
qualitative analysis and NLP. Their novelty lies in being
operationalized together as a single, inspectable pipeline for
corpus‑level reading that prioritizes coverage, traceability, and
self‑correction under scale.

\section{Results}\label{results}

\subsection{Evaluation Status}\label{evaluation-status}

The results from ReadingMachine should be understood as a demonstration
of system behavior at scale, rather than a formal evaluation of
performance. To date, evaluation consists of a single large-scale corpus
run and internal qualitative review.

The system was applied to a
\href{https://github.com/morrisseyj/ReadingMachine/blob/main/evaluation/industrial_policy_main_run/corpus.md}{corpus}
of 152 documents on industrial policy. The corpus is heterogeneous,
including academic papers, institutional reports (including long-form
documents exceeding 400 pages), books, and web-based materials. The
analysis was guided by research questions developed through Oxfam
America's formal research processes, reflecting real organizational
priorities.

The resulting output was reviewed internally by research staff at Oxfam
America. The outputs were assessed as substantively coherent and broadly
aligned with existing domain understanding. This review indicates that
the system produces plausible and interpretable representations of the
corpus. However, it does not constitute a systematic or independent
evaluation of performance.

Formal evaluation remains an open area of work. Key directions include:

\begin{itemize}
\tightlist
\item
  Expert review of outputs by domain specialists (e.g., macroeconomists
  and political economists working on international development)
\item
  Application across additional domains and corpora, with corresponding
  expert assessment
\item
  Comparative analysis against retrieval-augmented generation (RAG),
  hierarchical summarization pipelines, and off-the-shelf AI tools
  (noting that outputs differ in structure - see below - and require
  appropriate evaluation frameworks)
\item
  Formal benchmarking of system behavior
\item
  Evaluation of the practical usefulness of the outputs
\end{itemize}

To support this process, this paper constitutes an open invitation to
domain experts to
\href{https://github.com/morrisseyj/ReadingMachine/blob/main/evaluation/EVAL_INSTRUCTIONS.md}{review
the industrial policy output}. Reviews may be submitted via a structured
evaluation form and will be published in the project's GitHub repository
as part of an open review archive. Further collaboration on systematic
evaluation, including cross-domain testing and benchmarking, is actively
invited.

\subsection{Experimental Setup}\label{experimental-setup}

The corpus was constructed using the ReadingMachine *getlit* module, 
which combines structured search, LLM-assisted retrieval of grey literature, 
and human filtering. This produced a corpus of 175 documents, of which 152 
returned insights. This corpus is intentionally demanding: it is large, spans 
a diverse literature, and includes multiple document types, including webpages, 
academic papers, long institutional reports, and books. The final output included 
137 citations.

The analysis was guided by the following research questions:

\begin{itemize}
\tightlist
\item
  What drivers account for the resurgence of industrial policy in both
  highly industrial and industrializing countries?
\item
  How have the definitions and approaches to industrial policy shifted
  from the post-World War II period to the present, particularly with
  respect to sustainability, equality, and human rights?
\item
  What challenges and constraints do less-industrialized countries face
  in implementing effective industrial policy?
\item
  What recommendations can be made to industrialized countries to reduce
  harm to less-industrialized countries?
\item
  What reforms to transnational institutions could expand policy space
  for less-industrialized countries?
\end{itemize}

The pipeline was also provided with contextual framing derived from
Oxfam's Terms of Reference for the research. This framing informed both
document retrieval and insight extraction, ensuring that the corpus and
subsequent analysis were aligned with the intended scope of inquiry.

\subsection{System Metrics}\label{system-metrics}

The following metrics describe system behavior observed in a single
large-scale run. They are presented as descriptive indicators of
execution and output structure, rather than as measures of performance.

The full pipeline execution required approximately 14 hours and incurred
a total cost of approximately \$300.

Core system metrics include:

{\def\LTcaptype{none} 
\begin{longtable}[]{@{}
  >{\raggedright\arraybackslash}p{(\linewidth - 2\tabcolsep) * \real{0.4085}}
  >{\raggedright\arraybackslash}p{(\linewidth - 2\tabcolsep) * \real{0.5915}}@{}}
\toprule\noalign{}
\begin{minipage}[b]{\linewidth}\raggedright
Variable
\end{minipage} & \begin{minipage}[b]{\linewidth}\raggedright
Value
\end{minipage} \\
\midrule\noalign{}
\endhead
\bottomrule\noalign{}
\endlastfoot
Document count & 152 (175) \\
Final citation count & 137 \\
Total run cost & \textasciitilde{} \$300 \\
Insight count & 17752 \\
Chunk insight count & 12166 \\
Meta insight count & 5633 \\
Cluster count & 169 \\
Runs to stable schema & 5 \\
Theme counts & 34, 34, 34, 34, 34 \\
Orphan counts & 14008, 15092, 15318, 14927, 15657 \\
Failed theme counts & 4, 2, 2, 1, 0 \\
\end{longtable}
}

\subsection{Example Insights}\label{example-insights}

The following examples illustrate the form and level of granularity of
extracted insights. They are drawn from the output of the system.

{\def\LTcaptype{none}
\begin{longtable}[]{@{}
  >{\raggedright\arraybackslash}p{0.28\linewidth}
  >{\raggedright\arraybackslash}p{0.14\linewidth}
  >{\raggedright\arraybackslash}p{0.52\linewidth}@{}}
\toprule\noalign{}
\begin{minipage}[b]{\linewidth}\raggedright
Question
\end{minipage} & \begin{minipage}[b]{\linewidth}\raggedright
Insight type
\end{minipage} & \begin{minipage}[b]{\linewidth}\raggedright
Insight
\end{minipage} \\
\midrule\noalign{}
\endhead
\bottomrule\noalign{}
\endlastfoot
What challenges and constraints do less-industrialized countries face in
implementing effective industrial policy? & Chunk insight & `Many
developing countries had been busy dismantling their industrial policies
during the 1980s and the 1990s, which poses a challenge in realizing
effective industrial policy now (Chang et al. 2020).' \\
What drivers account for the resurgence of industrial policy in both
highly industrial and industrializing countries? & Meta insight & `The
resurgence of industrial policy in both highly industrialized and
industrializing countries is driven by the dual objectives of
technological advancement and environmental sustainability, as seen in
China's push for electric vehicles to enhance competitiveness and reduce
urban air pollution (Altenburg et al. 2017).' \\
\end{longtable}
}

\subsection{Qualitative
Observations}\label{qualitative-observations}

The
\href{https://github.com/morrisseyj/ReadingMachine/blob/main/evaluation/industrial_policy_main_run/industrial_policy_current.md}{output}
produced by the system is notable for its scale and density; readers are 
encouraged to inspect it directly. Even at a glance, it differs from 
conventional AI-generated summaries in ways that are immediately apparent:

\begin{itemize}
\tightlist
\item
  \textbf{Length}: The document is substantially longer than typical
  AI-generated outputs, often extending to tens of thousands of words.
\item
  \textbf{Density}: Sections are tightly packed with discrete claims,
  with relatively little compression into high-level generalizations.
\item
  \textbf{Granularity}: Specific mechanisms, conditions, and
  qualifications are retained rather than absorbed into broader
  summaries.
\item
  \textbf{Persistence of detail}: The output does not quickly converge
  to a concise narrative; instead, it maintains a large number of
  localized statements drawn from across the corpus.
\item
  \textbf{Coherence under scale}: Despite the volume of detail, sections
  read as continuous, structured prose rather than fragmented or
  list-like outputs.
\item
  \textbf{Breadth without compression}: The document remains readable
  while operating at a level of coverage and detail that is atypical of
  standard summarization or retrieval-based workflows.
\end{itemize}

Taken together, these properties give the output a character that is
immediately distinguishable from query-based or summarization-driven
systems, which typically produce shorter, more compressed
representations. The output reads more like an academic review of the
literature.

Looking more closely at the output, the system produces a structured
mapping of the corpus rather than a single narrative argument. The
separation of reading (insight extraction and organization) from
downstream interpretation is visible in the descriptive character of the
text, which represents the literature without advancing a synthesized
position.

Based on internal review, the themes generated appear interpretable 
and broadly consistent with recognizable structures in the industrial policy literature. 
No themes were identified as clearly distorted or
artificial. The overall output reads more like an academic literature review than 
a conventional LLM summary or question-answering response, 
and remains coherent across research questions. The re-theming process appears 
to consolidate conceptual structure across iterations: the reduction in theme count between passes
is consistent with initial semantic groupings being reorganized into
more coherent thematic categories.

Granular claims appear to be preserved throughout the output. The system does not
collapse content into high-level abstractions; instead, specific and
detailed claims remain visible within thematic summaries, contributing
to the overall density of the document.

Conflict is explicitly represented. Tensions within the literature are
articulated and are not systematically smoothed over during synthesis.

No obvious hallucinations were identified during internal review, though
this has not been formally or systematically evaluated.

Notably theme count remains stable across iterations. In previous 
testing runs, themes initially split to address output constraints 
(though the starting number of themes was fewer in these cases) and 
then stabilized. Nonetheless the total number of failing themes declines 
across runs indicating that the iterative approach is working to address 
output constraints, while maintaining high levels of granularity and 
limiting omission.

The distribution of content across themes is uneven. Some themes are
significantly more developed than others. This likely reflects variation
in the density of the underlying literature, rather than a constraint
imposed by the system. In contrast, human-led reviews often impose
balance across sections, even where this diverges from the distribution
of available evidence.

The pipeline produces an output that is structurally distinct from conventional 
AI systems, including query-conditioned and summarization-based approaches. 
Direct comparison with systems such as RAG is therefore not straightforward, 
as the outputs differ in form and objective; even where comparisons are possible 
(e.g., coverage or traceability), the central evaluation question is whether 
separating reading from reasoning in this way leads to better downstream reasoning, 
sufficient to justify the additional cost, latency, and complexity.

\subsection{Failure Modes and
Issues}\label{failure-modes-and-issues}

\subsubsection{Large proportion of
meta-insights}\label{large-proportion-of-meta-insights}

The meta-insight generation step was the most computationally expensive
stage of the pipeline, accounting for approximately \$180 of total cost
and producing roughly one third of all insights. This is notable, as the
meta pass was originally conceived as a supplement to chunk-level
extraction, intended to capture only those arguments that emerge across
extended portions of a document.

Given the observed volume of meta-insights, it is likely that the
meta-insight prompt is overly permissive. Notably, an earlier run
produced higher meta-insight:insight proportions (\textasciitilde1:2),
reflecting a too loose meta-insight extraction prompt and a too
restrictive chunk extraction prompt. Overall this prior run produced
more compressed and narratively coherent output, but with less granular
representation. The current run relaxed instructions for chunk-level
extraction, and tightened those for meta-insight extraction. The result
was better coverage at the cost of redundancy and increased
computational pressure. This suggests two important reflections: 1) the
system presents as parameterizable where different configurations can
produce different trade-offs between compression, coverage, and
scalability. 2) Prompt specification likely plays a significant role in
controlling insight generation. This allows for system tuning, but might
make model swapping more complicated, and raises the possibility that
simultaneously tuning meta-insight extraction and chunk insight
extraction to complementary points may be non-trivial. For the purposes
of this release, extensive tuning of this component was not pursued.

The prevalence of meta-insights also introduces potential challenges for
reproducibility. Because meta-insights are derived from near-full
document context, they are likely to be less stable across runs than
chunk-level insights. Variability at this stage may propagate through
clustering and theme generation, reducing structural consistency. This
effect may be limited when meta-insights constitute a small proportion
of the total, but becomes more significant when they dominate the
insight set.

A further contributing factor may be the model's ability to track
previously extracted insights under context window pressure. If prior
chunk-level insights are not effectively retained, the model may
reproduce or recombine existing claims during the meta pass, leading to
redundancy. Under this interpretation, the meta-insight set may include
a mixture of previously extracted chunk-level claims, recombinations of
such claims, and genuinely cross-cutting arguments.

These factors together suggest that the observed volume of meta-insights
reflects both prompt specification and context management limitations.
The resulting redundancy appears to be partially resolved during
downstream synthesis, but increases computational cost and may affect
reproducibility. Refinement of the meta-insight extraction process
remains a priority for future work.

\subsubsection{High orphan counts}\label{high-orphan-counts}

Orphan counts remain high across all passes and in fact show an upward trend. 
This is partly expected for two reasons. First, the orphan identification prompt 
is intentionally strict in order to prioritize coverage: false positives are preferred 
to missed insights. Second, summaries are fully regenerated during the theme population 
step rather than incrementally updated with previously reinserted orphans. As a result, 
improvements introduced during orphan handling are not directly preserved in subsequent 
synthesis steps. If the context window is under pressure during theme population, and 
insights are dropped at that stage, this behavior is likely to persist even as the theme 
schema improves.

Taken together, these effects suggest that the model is operating under
synthesis pressure. This likely reflects both the size of the input and
the heterogeneity of the insight set, which can lead the model to
produce increasingly abstract summaries. Under these conditions,
individual claims may be incorporated at a conceptual level without
remaining directly recoverable in near-verbatim form. In such cases, the
orphan detection step may classify insights as unrepresented even when
they have been integrated at a higher level of abstraction.

These dynamics indicate that high orphan counts reflect the interaction
between conservative omission detection and abstraction during
synthesis. In contrast to conventional summarization workflows---where
information loss is often silent and difficult to observe---the orphan
mechanism makes such loss explicit, allowing it to be detected and
addressed rather than assumed away.

This behavior also suggests a broader reinterpretation of the synthesis
process implemented by ReadingMachine. The system was initially designed
around delayed compression: preserving high-granularity representations
until thematic structure was established, with orphan handling acting as
a safeguard against omission. However, the observed scale of orphan
reinsertion indicates that the system is not only compressing late, but
also performing substantial late-stage reinflation, in which underlying
insights are reintroduced after abstraction.

This contrasts with most synthesis pipelines, where compression is
assumed to be monotonic and information flow strictly reductive. Under
that assumption, sufficiently capable models are expected to retain all
salient information during summarization. The behavior observed here
suggests a different dynamic: synthesis may instead involve an iterative
process in which abstraction is repeatedly challenged by the
reintroduction of underlying material. Under explicit coverage
constraints, maintaining representational fidelity may require adding
information back in, rather than assuming that abstraction alone
preserves substance.

One interpretation of this behavior is that language models do not
explicitly track which claims have been preserved, abstracted, or
omitted during synthesis. They cannot therefore verify coverage through
generation alone. The orphan handling mechanism accounts for this by
externalizing this process, treating coverage as something that must be
tracked and enforced across coordinated steps rather than inferred from
fluent output.

\subsubsection{Compute and run time}\label{compute-and-run-time}

The computational cost and runtime of the system are non-trivial, and
their significance depends on the outcome of future evaluation. If
ReadingMachine provides only marginal improvements over lower-cost AI
approaches (see complementarity below), these costs would represent a
substantial limitation.

However, if the system addresses meaningful failure modes in existing
approaches and contributes to the research workflow, the more
appropriate comparison is human-led review. In that context, the system
represents a reduction in time and cost of one to two orders of
magnitude. Under this framing, the relative importance of computational
inefficiencies---such as the cost of the meta-insight generation
step---becomes less significant.

\subsubsection{Claim tracing}\label{claim-tracing}

Insights are currently referenced using an author--date--year format. As
a result, recall operates at the level of themes and source documents
rather than individual claims. In cases where multiple documents
contribute to a theme, tracing an insight may surface several relevant
chunks (and meta-level segments), making it more difficult to locate the
precise originating text.

This limitation could be addressed in future iterations by linking each
insight\_id directly to its source span, with human-readable
author--date references layered on top. The implementation of such
functionality is dependent on the output format---for example, HTML
hover interactions versus jump links in PDF or Word documents.

For meta-insights in particular, tracing may return large portions of
text even when a single insight\_id is queried. This reduces the
precision of traceability and reinforces the need to further refine the
meta-insight extraction process.

\subsubsection{Dropped citations}\label{dropped-citations}

As mentioned above the dropping of citations that are known to 
hold insights is a concern for the method. It raises the questions not 
only about citation provenance but also, more concerningly, about whether 
the insights themselves are being lost. Given the orphan insertion loop, 
it is thought unlikely that there is significant insight loss and that this 
is a citation tracking/model bias towards coherence issue, but this requires 
evaluation. It is quite possible that citation repair is another process of 
orphan handling. 

Addressing the issue of dropped citations therefore raises three future areas of work:
\begin{itemize}
  \item Under the current implementation add fuzzy string matches or semantic matches to handle 
deterministic string replacement in the application of patch repair.
  \item Evaluate insight coverage to check whether the orphan loop is working sufficiently 
and identify whether dropped citations are simply an issue of coherence bias/tracking
  \item Test an approach to orphan handling which mimics the schema and citation repair 
process which removes the coherence requirement from the model when it comes to inserting 
orphans.
\end{itemize}

For item 3 above, the proposed change in approach is not obviously better as this task sits 
on the boundary of synthesizing information (the orphan) and tracking that you have addressed 
it. Moving this to more of a tracking framework may result in a less coherent output. 

\subsection{Interpretation and Next Steps}\label{interpretation-and-next-steps}

Based on the observations above, the architecture appears to function as
intended. It produces structured, traceable representations of large
corpora, preserves granular claims, and maintains internal coherence
without obvious hallucination. Fundamentally, the output looks different
to both human-produced reviews and other AI approaches, in ways that
reflect its architecture.

However, the extent to which this approach improves upon and/or augments
existing methods remains an open empirical question. Addressing this
question requires systematic evaluation, including expert review,
cross-domain testing, and formal comparison with alternative approaches.

The current results should therefore be understood as a demonstration of
system behavior at scale, rather than as a definitive assessment of
performance.

\section{Implications}\label{implications}

If the observed system behavior generalizes beyond this initial run,
ReadingMachine has implications both for the practice of large-scale
synthesis and for how knowledge is produced and evaluated in
institutional settings. These potential implications are discussed
below.

\subsection{Cost and Time
Reductions}\label{cost-and-time-reductions}

By shifting the labor of large-scale reading from human analysts to
structured machine processes, the system substantially reduces both the
cost and time required to synthesize large corpora. Tasks that would
typically require months of expert review can be completed in hours at a
fraction of the cost. While computational costs are non-trivial, they
remain orders of magnitude lower than comparable human-led efforts,
particularly as corpus size increases.

\subsection{A Shift in the Bottleneck of Knowledge
Work}\label{a-shift-in-the-bottleneck-of-knowledge-work}

As the cost of reading declines, the primary constraint in knowledge
production shifts. For institutions, the bottleneck moves away from
information acquisition and toward:

\begin{itemize}
\tightlist
\item
  absorption and internalization of findings\\
\item
  organizational learning\\
\item
  updating prior assumptions\\
\item
  decision-making and strategic pivoting
\end{itemize}

In this sense, ReadingMachine does not eliminate human labor; it
reallocates it from reading to interpretation and action.

\subsection{Increased Coverage}\label{increased-coverage}

Because the system performs a structured pass over the entire corpus, it
reduces the risk of partial engagement that characterizes both human
review and retrieval-based AI systems. All documents are processed, and
insights are extracted systematically. Mechanisms such as orphan
detection further reinforce coverage by identifying and reintegrating
omitted material.

\subsection{Inspectability and
Traceability}\label{inspectability-and-traceability}

Each stage of the pipeline produces intermediate artifacts---chunks,
insights, clusters, themes---that can be inspected directly. This makes
it possible to trace any element of the final synthesis back to its
source material. In contrast to both human synthesis (where intermediate
reasoning is rarely visible) and many AI systems (where transformations
are opaque), ReadingMachine exposes the structure of the analytical
process.

\subsection{Separation of Reading and
Reasoning}\label{separation-of-reading-and-reasoning}

A central implication of the ReadingMachine approach is the explicit
separation of reading from reasoning. In most forms of human synthesis,
these processes are intertwined: analysts read, interpret, evaluate, and
draw conclusions simultaneously, often in ways that are difficult to
observe or reconstruct. This coupling was historically necessary when
reading was slow, as judgment guided which arguments to engage with, how
to resolve debates and whether to balance narrative weights. However,
this muddies claims of what the literature says with what the author
thinks of the literature, and therefore limits the transparency of how
conclusions are formed.

ReadingMachine decomposes this process. It constrains language models to
perform bounded reading tasks---such as extracting and organizing
claims---while deferring interpretation, evaluation, and judgment to the
researcher or to downstream analytical steps. The system therefore
produces a structured representation of what the corpus contains, rather
than an answer to what it means or what should be concluded.

This separation has two implications. First, it increases transparency:
the intermediate representations make it possible to examine how the
corpus has been read before any higher-level reasoning is applied.
Second, it allows reasoning to be applied more deliberately and, if
necessary, iteratively, using the structured output as a stable input.
In this sense, the system replaces a single, opaque synthesis step with
a sequence of inspectable transformations followed by explicit
interpretation.

It is important to note that reading itself remains inherently
interpretive. Both human readers and language models rely on prior
frameworks (conceptual in the case of humans, and learned semantic
patterns in the case of model) to identify and express what counts as an
insight. ReadingMachine does not eliminate this interpretive layer.
Instead, it makes these transformations visible by externalizing
intermediate outputs. While reasoning over the corpus is deferred,
interpretive judgments made during reading are exposed and can be
examined, compared, and, where necessary, challenged. This creates
opportunities for epistemic analysis, discussed further below.

\subsection{Reproducibility}\label{reproducibility}

Reproducibility in this system is conceptual rather than semantic.
Re-running the pipeline with the same configuration is not expected to
produce identical text, but it should produce a similar thematic
structure: comparable themes, consistent groupings of insights, and
stable relationships between them.

This form of reproducibility is achieved through:

\begin{itemize}
\tightlist
\item
  decomposition of the reading process into constrained, repeatable
  steps
\item
  use of structured intermediate representations (insights, clusters,
  themes)
\item
  fixed analytical configurations (e.g., prompts, models, clustering
  parameters)
\end{itemize}

Under these conditions, variation across runs is expected to remain at
the level of phrasing and emphasis rather than underlying structure. At
larger scales, this stability is expected to degrade as clustering
noise, context inputs, and corpus heterogeneity increase. This serves to
reduce reproducibility. These scaling effects are discussed further in
the limitations section.

\subsection{Architectural Scaling}\label{architectural-scaling}

ReadingMachine is designed to operate on corpora substantially larger
than those typically handled by retrieval systems or hierarchical
summarization workflows. However, it is not assumed to be infinitely
scalable, and its behavior at scale should be understood as an open
empirical question.

The key architectural difference is where scale constraints are expected
to appear. In many retrieval or agentic workflows, the context
bottleneck occurs during the final synthesis step, when the model must
integrate all relevant information gathered across the corpus. As the
amount of material grows, this can stress the context window and lead to
known failure modes such as attention loss or ``missing middle''
effects, where portions of the input receive less effective processing.
This increases omission risk and complicates citation anchoring.

ReadingMachine is designed to shift this limitation by converting
documents into a structured intermediate representation before
synthesis, distributing integration across smaller, theme-level
synthesis tasks rather than concentrating it in a single corpus-level
step. Those intermediate representations are as follows:

documents $\rightarrow$ chunks $\rightarrow$ insights $\rightarrow$ clusters $\rightarrow$ themes $\rightarrow$ synthesis

Because synthesis occurs at the theme level rather than the corpus
level, the model does not need to comprehend the entire corpus. Instead,
it integrates the insights associated with one thematic area at a time.
The only stage where broader integration occurs is theme schema
generation, which is scaffolded by cluster summaries or prior theme
structures and repeated iteratively.

In theory, this decomposition should allow the system to scale beyond
the point at which corpus-level synthesis becomes unstable in other
approaches.

\subsection{Parameterization and Experimental
Control}\label{parameterization-and-experimental-control}

Because each stage of the pipeline is explicitly defined, the system can
be treated as an experimental framework. Researchers can hold most
parameters constant while varying individual components---such as
models, prompts, or clustering settings---to observe how these changes
affect the resulting thematic structure. This enables systematic
sensitivity analysis that is difficult to achieve in both human-led
synthesis and less structured AI workflows. Note however, that such
experimentation requires system stability, and therefore
reproducibility. As mentioned above (and discussed below) it is not
clear this will hold as scale increases.

\subsection{Implications for Epistemic
Analysis}\label{implications-for-epistemic-analysis}

As noted above, beyond synthesis, ReadingMachine creates opportunities
for epistemic experimentation in understanding how knowledge is
constructed. A full exploration of these epistemic implications is
beyond the scope of this paper, but the system provides a structured
basis for such analysis. By comparing outputs across models or
configurations, researchers can observe how different systems organize,
prioritize, or omit information. These differences can be interpreted as
expressions of underlying assumptions or priors embedded in models or
analytical choices.

In this sense, the system can be used not only to map a corpus, but also
to study the process of mapping itself---providing a structured basis
for critical analysis of how interpretations are formed. Under this
approach, differences across model runs expose model biases, not as
errors requiring correction, but as signals that provide insight into
how different systems structure and represent knowledge. Rather than
converging on a single ``correct'' interpretation, the method enables
comparison across alternative, internally coherent representations of
the same material.

Together, these properties suggest a shift from selective, narrative
synthesis toward structured, inspectable corpus mapping as a distinct
mode of knowledge production.

\subsection{The need for external LLM auditing}\label{the-need-for-external-llm-auditing}

The quantity of external structure required for ReadingMachine to 
maintain coverage, granularity, citation provenance, and thematic 
coherence under scale suggests that unconstrained LLM synthesis 
does not naturally conserve these properties. Throughout development, 
multiple forms of representational drift were observed, including 
dropped insights, disappearing citations, unstable theme structures, 
and incomplete repairs. Notably, these failures emerged not during 
complex reasoning tasks, but during comparatively bounded comprehension, 
synthesis, and rewriting operations.

In this respect, ReadingMachine can be understood in part as an external 
auditing architecture for iterative LLM synthesis. Mechanisms such as 
schema development, insight mapping, orphan handling and citation 
verification act as external conservation controls, ensuring that coverage, 
provenance, and structural fidelity are not silently lost during repeated 
generative transformations.

Notably, the auditing requirements observed in ReadingMachine are largely 
orthogonal to current dominant scaling strategies in LLM development, 
such as context-window expansion. The behavior observed in the development 
of ReadingMachine suggests that increasing context windows may increase 
rather than eliminate the need for externalized auditing. Larger synthesis 
contexts likely amplify omission pressure, abstraction pressure, citation 
drift, and instability in maintaining coherent representational structure 
across repeated transformations. In this respect, visibility over larger 
amounts of text should not be conflated with stable integration of that 
text into a faithful synthesis.

While the instability documented here concerns synthesis and comprehension 
tasks, the same underlying generative dynamics are likely relevant to 
reasoning processes, which also operate through iterative transformations 
over bounded texts. To the extent that reasoning depends on preserving 
assumptions, constraints, provenance, or intermediate conclusions across 
extended inference trajectories, similar forms of representational drift 
may emerge. This suggests that long-horizon reasoning systems may also benefit 
from external auditing architectures in cases where completeness, consistency, 
traceability, or preservation of epistemic structure are important.

\subsection{Fighting upstream compression}\label{fighting-upstream-compression}

A core element of ReadingMachine's design was to delay compression as long as 
possible. However even with that in mind, a repeated pattern appeared in 
building the workflow. The model would compress outputs, to the detriment 
of coverage, even when explicitly instructed not to. This was apparent in 
orphan insertion, which was initially conceived of as defensive bootstrapping 
but which turned out to be necessary reinflation of compressed material. 
Likewise with schema repair, the model would anchor on conceptually elegant 
schema, going so far as to hallucinate repairs. And similarly with citations, 
the model would simply fail to integrate citations due to its apparent anchoring 
on more elegant prose. 

One interpretation is that the model is optimized more strongly for producing 
coherent synthesized outputs than for conserving all source information through 
iterative transformations. This suggests that for a system like ReadingMachine, 
the operational pattern is to put in place auditing architecture like that 
described above and separate out repair tasks from those seeking to synthesize 
information. Effectively fight information loss via compression until completeness 
is achieved, and only then allow some compression. 

It is worth noting that the above failures were not a result of context pressure. 
When we provided the exact same context but with instructions that ignored synthesis - 
generate a repair plan, or generate patches - results improved enormously. This again 
points to the idea that the solutions to effective information preserving synthesis 
might not be larger context windows - which could actually make the problem worse 
under models with similar training imperatives. They might be auditing and task separation.

More generally, the repeated success of decomposition, patch generation, auditing, 
and repair suggests that information-preserving workflows may benefit from separating 
representational fidelity tasks from synthesis tasks, rather than assuming that a 
single generative operation can optimize both simultaneously. 

\section{Limitations}\label{limitations}

ReadingMachine is designed for a specific analytical task: high-fidelity
thematic synthesis of large natural-language corpora. It is not intended
to replace other approaches to working with language models, nor should
it be considered universally superior. In many contexts---particularly
exploratory analysis or question answering---other methods remain more
appropriate (see below for complementarity with other approaches).

\subsection{Cost}\label{cost}

Assuming the results above generalize and are validated, ReadingMachine,
stands to reduce the cost of human-led synthesis by one to two orders of
magnitude. That said, the pipeline is not costless. The system performs
multiple passes over the entire corpus, incurring both computational and
time costs. As a result, it is less well suited to rapid exploration or
ad hoc query answering.

Most components of the pipeline scale linearly and predictably with
corpus size. However, certain stages---particularly those involving
iterative synthesis and reinsertion---can introduce non-linear scaling
effects, even if these are not the dominant contributors to total cost.

There are clear pathways for reducing cost. The most immediate is
parallelization of language model calls, which are currently executed
sequentially. The pipeline is well suited to parallelization, as its
most expensive stage---insight extraction---operates independently
across chunks and documents.

Additional efficiencies may be achieved through model specialization.
Smaller models could be used for bounded tasks such as insight
extraction, though higher-capability models are currently preferred
given the importance of insight quality in downstream stages. Over time,
the system may enable a data flywheel: high-quality, structured insight
outputs can serve as training data for smaller, task-specific models,
reducing reliance on more expensive general-purpose models.

\subsection{Functional Scope}\label{functional-scope}

The pipeline does not reason over the corpus. It cannot answer questions
such as ``how many studies did x'' or ``what is the strongest method for
examining y'' unless such claims are explicitly present in the source
material. Nor does it independently assess the credibility or weight of
claims unless this is discussed within the corpus itself. Its output is
therefore a structured mapping of what the corpus contains, not an
evaluation of what should be concluded.

This places greater importance on question formulation and corpus
selection, which are left entirely to the user. These upstream decisions
shape the scope and relevance of the resulting synthesis. Because the
system faithfully maps the corpus it is given, biases, omissions, or
imbalances in corpus selection will be reflected in the output.

While this constraint limits the range of tasks the system can perform,
it also underpins several of its advantages. By avoiding open-ended
reasoning, the pipeline increases traceability, inspectability, and
reproducibility. Rather than embedding interpretation within a single
opaque step, it relocates reasoning to distinct stages: upstream
(question definition and corpus selection), external (parameter
choices), and downstream (interpretation and decision-making).

\subsection{Performance Under Scale}\label{performance-under-scale}

The primary scaling constraints in ReadingMachine arise from two
interacting factors: the heterogeneity of insights, which affects the
stability of intermediate representations, and the volume of insights
assigned to individual themes, which affects the system's ability to
integrate them during synthesis. These constraints emerge at different
stages of the pipeline and produce distinct but interacting failure
modes.

\subsubsection{Representation Under
Heterogeneity}\label{representation-under-heterogeneity}

As corpus heterogeneity increases, the semantic structure of the insight
space becomes more difficult to represent consistently. Small variations
in embeddings can lead to different clustering outcomes, particularly as
the number and diversity of insights grow. Because clusters serve as
scaffolding for theme generation, this variability propagates into the
theme schema, altering how the corpus is organized conceptually.

These effects reduce system stability across runs. Even with identical
configurations, small stochastic differences in model outputs can
produce different cluster structures and, in turn, different thematic
organizations. More heterogeneous corpora are therefore expected to
yield less stable representations and may require additional iterations
to converge on a coherent structure.

\subsubsection{Synthesis Under Volume}\label{synthesis-under-volume}

The mechanisms introduced in the pipeline to handle scale significantly
extend the size of corpora that can be processed relative to other
AI-supported synthesis methods. However, they do not eliminate scaling
constraints entirely. At sufficiently large scales, the primary
limitation is expected to arise during theme schema generation, when
cluster summaries may become too large to fit within the model's context
window.

One possible approach is to provide the schema generation step with only
a partial view of the underlying content and rely on the orphan handling
and iterative re-theming process to recover omitted material. However,
the effectiveness of this approach remains uncertain and requires
further empirical evaluation.

\subsubsection{Reproducibility under
scale}\label{reproducibility-under-scale}

A fundamental limitation of the system is that language model
determinism diminishes as context windows increase. As a result, several
central stages of the pipeline---particularly those that rely on large
context windows, such as meta-insight generation, cluster summarization,
and theme population---are inherently less stable. As scale increases,
pressure on the context window grows, further reducing determinism and
increasing variability across runs. This suggests that the system's
structural stability is likely to decline at larger scales. The current
high proportion of meta-insights may exacerbate this effect, reinforcing
the importance of refining the meta-insight extraction process.

It is important to note that reduced stability does not necessarily
imply reduced accuracy. A given corpus may admit multiple valid thematic
organizations, each internally coherent but differing in structure. In
this sense, variation across runs may reflect alternative, but
plausible, representations rather than outright error. The system is
therefore expected to remain useful at scale for producing structured
outputs, even if exact reproducibility is limited.

However, this instability has implications for experimental use. While
large-scale runs may produce useful and interpretable outputs,
comparative analysis across runs---such as evaluating the effects of
parameter changes---may become less reliable as scale increases. This
suggests two modes of use: large-scale execution for generating
structured syntheses, and smaller-scale runs for controlled
experimentation and analysis.

\subsection{Coverage and Redundancy Trade-offs}\label{coverage-and-redundancy-trade-offs}

A central aim of ReadingMachine is to maximize coverage and minimize
omission. While the system is designed to approach this
asymptotically---through full-corpus processing and orphan
reinsertion---it cannot guarantee complete coverage, particularly at
larger scales.

As volume increases, the system prioritizes local completeness within
individual themes. Because themes are synthesized independently, this
can lead to redundancy across the final output. This redundancy is an
intentional trade-off: it preserves marginal or low-salience insights at
the cost of repetition. As scale increases further, both omission risk
and redundancy pressure tend to rise simultaneously.

These limitations reflect trade-offs inherent to the system's design.
Prioritizing coverage, traceability, and inspectability shifts
constraints toward scale, stability, and computational cost.

\section{Complementarity with Other AI Approaches}\label{complementarity-with-other-ai-approaches}

ReadingMachine differs from both hierarchical summarization and
retrieval-based systems. Unlike hierarchical summarization, it delays
compression rather than aggregating it through successive summaries.
Unlike retrieval-augmented generation (RAG) and its agentic variants, it
performs a structured pass over the entire corpus rather than sampling
subsets based on queries.

These differences allow ReadingMachine (assuming preliminary results are
validated and generalize) to address limitations in existing
approaches---particularly silent omission and early information loss.
This makes it particularly suitable to work in research, policy
analysis, legal/regulatory review, and efforts at institutional memory.
That said, the design objectives also result in specific constraints.
ReadingMachine is therefore not a replacement for these methods, but a
complementary approach designed for a distinct analytical task.

RAG and agentic systems are well suited to exploratory workflows. They
are fast, flexible, and effective when questions are evolving, partial
coverage is acceptable, and low latency is important. ReadingMachine is
better suited to later stages of research, when questions are more
stable and the goal is structured synthesis. It is most useful when
omission is costly, edge cases must be preserved, and a comprehensive
mapping of the corpus is required.

These approaches can be combined within a broader research workflow:

\begin{itemize}
\tightlist
\item
  Agentic research systems support early exploration and corpus
  identification
\item
  RAG enables flexible, query-driven exploration of the corpus
\item
  ReadingMachine performs a structured reading pass, producing a
  thematic map of arguments and insights
\item
  Agentic or retrieval workflows can then be used to interrogate the
  resulting structure, explore edge cases, and support further analysis
\end{itemize}

A typical workflow might therefore proceed as follows:

\begin{enumerate}
\item Agentic search $\rightarrow$ corpus identification
\item RAG exploration $\rightarrow$ initial familiarity with the literature
\item ReadingMachine $\rightarrow$ structural mapping of arguments and themes
\item Agentic or retrieval workflows $\rightarrow$ deeper analysis and follow-up questions
\end{enumerate}

\section{Status of ReadingMachine and Future
Work}\label{status-of-readingmachine-and-future-work}

ReadingMachine is currently an early-stage experimental method and
open-source implementation
(\href{https://github.com/morrisseyj/ReadingMachine/}{GitHub repository}).
The project is intended as a methodological framework as much as a software 
tool, and its robustness, limitations, and boundary conditions require empirical
evaluation across domains.

The project is actively seeking collaborators to apply the system in
different contexts, using varied corpora and configurations. Particular
emphasis is placed on identifying failure modes, assessing thematic
quality, and evaluating the stability of outputs under different
conditions. Contributions that challenge or stress the pipeline are
especially valuable.

Active collaboration is also being sought for code hardening and feature
development. Priority areas include:

\begin{itemize}
\tightlist
\item
  Parallelization
\item
  Benchmarking
\item
  Tuning
\item
  Citation management and provenance
\item
  OCR implementation
\item
  Model interchangeability
\item
  Documentation improvement
\item
  Strategic and methodological reflection
\end{itemize}

A potential direction for future work is the introduction of an explicit
downstream reasoning stage operating over the structured outputs
produced by ReadingMachine. Rather than performing unconstrained
reasoning directly over raw corpora, such a stage would apply
interpretive analysis to a stable, traceable representation of the
material. This could involve agentic workflows that incorporate external
tools (e.g., web search or retrieval systems built on the underlying
corpus), while requiring explicit justification of claims, articulation
of trade-offs, and transparency in how conclusions are formed.

Under this approach, reasoning would be applied in a controlled and
inspectable manner, potentially operating over individual themes or
structured subsets of the corpus. Different interpretive stances (e.g.,
distribution-sensitive, precautionary, growth-oriented) could be applied
explicitly, allowing alternative readings of the same material to be
compared. In this sense, such a layer would not replace the mapping
function of ReadingMachine, but build on it---transforming structured
representations into interpretable analytical outputs through a distinct
and traceable step.

This direction remains exploratory. It is not assumed to yield canonical
answers, eliminate bias, or provide definitive conclusions. Rather, it
is proposed as a way to reintroduce reasoning into the synthesis process
in a manner that is explicit, inspectable, and consistent with the
methodological principles of ReadingMachine.

\section{Conclusion}\label{conclusion}

The growth in the volume of written information has created a structural
imbalance between what is available to be known and what can be
systematically read and synthesized. Existing approaches---both human
and computational---address this problem only partially, often
introducing forms of omission that are difficult to detect.

ReadingMachine proposes a different approach. By decomposing reading
into structured, inspectable tasks and coordinating language models to
execute them across an entire corpus, the system shifts the bottleneck
in knowledge work from information access to interpretation and
decision-making. In doing so, it enables more comprehensive, traceable,
and reproducible synthesis of large document collections, while
explicitly separating the act of reading from downstream reasoning.

The method does not eliminate interpretation, does not replace human
judgment, and does not supersede existing AI tools. Instead, it
complements these approaches by introducing a structured reading layer
that produces inspectable intermediate representations of the
corpus---representations that can be treated as explicit epistemic
objects, examined prior to interpretation. This reorganization allows
reasoning to be applied more deliberately and transparently, rather than
being embedded within a single opaque synthesis step.

As both a technical framework and a methodological proposal,
ReadingMachine represents a step toward industrial-scale reading that
preserves the structure and diversity of the underlying corpus.


\begin{thebibliography}{99}

\bibitem{he2025}
Pengfei He, Zhenwei Dai, Bing He, Hui Liu, Xianfeng Tang, Hanqing Lu, Juanhui Li, Jiayuan Ding, Subhabrata Mukherjee, and Suhang Wang.
\newblock TRAJECT-Bench: A trajectory-aware benchmark for evaluating agentic tool use.
\newblock \emph{arXiv preprint arXiv:2510.04550}, 2025.

\bibitem{laban2024}
Philippe Laban, Alexander Richard Fabbri, Caiming Xiong, and Chien-Sheng Wu.
\newblock Summary of a haystack: A challenge to long-context LLMs and RAG systems.
\newblock In \emph{Proceedings of EMNLP}, pages 9885--9903, 2024.

\bibitem{liu2024}
Nelson F. Liu, Kevin Lin, John Hewitt, Ashwin Paranjape, Michele Bevilacqua, Fabio Petroni, and Percy Liang.
\newblock Lost in the middle: How language models use long contexts.
\newblock \emph{Transactions of the Association for Computational Linguistics}, 12:157--173, 2024.

\bibitem{narayan2018}
Shashi Narayan, Shay B. Cohen, and Mirella Lapata.
\newblock Don't give me the details, just the summary! Topic-aware convolutional neural networks for extreme summarization.
\newblock In \emph{Proceedings of EMNLP}, pages 1797--1807, 2018.

\bibitem{lewis2020}
Patrick Lewis, Ethan Perez, Aleksandra Piktus, Fabio Petroni, Vladimir Karpukhin, Naman Goyal, Heinrich K{\"u}ttler, Mike Lewis, Wen-tau Yih, and Tim Rockt{\"a}schel.
\newblock Retrieval-augmented generation for knowledge-intensive NLP tasks.
\newblock In \emph{Advances in Neural Information Processing Systems}, volume 33, pages 9459--9474, 2020.

\bibitem{cohan2018}
Arman Cohan, Franck Dernoncourt, Doo Soon Kim, Trung Bui, Seokhwan Kim, Walter Chang, and Nazli Goharian.
\newblock A discourse-aware attention model for abstractive summarization of long documents.
\newblock In \emph{Proceedings of NAACL-HLT, Volume 2: Short Papers}, pages 615--621, 2018.

\bibitem{yao2022}
Shunyu Yao, Jeffrey Zhao, Dian Yu, Nan Du, Izhak Shafran, Karthik Narasimhan, and Yuan Cao.
\newblock ReAct: Synergizing reasoning and acting in language models.
\newblock \emph{arXiv preprint arXiv:2210.03629}, 2022.

\bibitem{arksey2005}
Hilary Arksey and Lisa O'Malley.
\newblock Scoping studies: Towards a methodological framework.
\newblock \emph{International Journal of Social Research Methodology}, 8(1):19--32, 2005.

\bibitem{braun2006}
Virginia Braun and Victoria Clarke.
\newblock Using thematic analysis in psychology.
\newblock \emph{Qualitative Research in Psychology}, 3(2):77--101, 2006.

\bibitem{grootendorst2022}
Maarten Grootendorst.
\newblock BERTopic: Neural topic modeling with a class-based TF-IDF procedure.
\newblock \emph{arXiv preprint arXiv:2203.05794}, 2022.

\bibitem{mcinnes2018}
Leland McInnes, John Healy, and James Melville.
\newblock UMAP: Uniform manifold approximation and projection for dimension reduction.
\newblock \emph{arXiv preprint arXiv:1802.03426}, 2018.

\bibitem{mcinnes2017}
Leland McInnes, John Healy, and Steve Astels.
\newblock HDBSCAN: Hierarchical density-based clustering.
\newblock \emph{Journal of Open Source Software}, 2(11):205, 2017.

\end{thebibliography}
\end{document}